\documentclass[letterpaper, 10 pt, conference]{ieeeconf}  

\IEEEoverridecommandlockouts                              

\usepackage{graphics} 
\usepackage{epsfig} 
\usepackage{mathptmx} 
\usepackage{times} 
\usepackage{amsmath} 
\usepackage{amssymb}  
\usepackage{xspace}
\usepackage{booktabs}
\usepackage{multirow}
\usepackage{threeparttable}
\usepackage{afterpage}
\usepackage{caption}
\usepackage{cuted}

\newcommand{\ie}{\emph{i.e.}\xspace}
\newcommand{\eg}{\emph{e.g.}\xspace}

\usepackage{cite}
\makeatletter
\let\NAT@parse\undefined
\makeatother
\usepackage[colorlinks=true,   
            linkcolor=blue,   
            citecolor=blue,   
            urlcolor=blue,    
            pdfborder={0 0 0} 
]{hyperref}

\title{\LARGE \bf
DepthVLA: Enhancing Vision-Language-Action Models with Depth-Aware Spatial Reasoning
}

\author{%
    Tianyuan Yuan$^{1,2}$, 
    Yicheng Liu$^{1,2}$,
    Chenhao Lu$^{1,2}$,
    Zhuoguang Chen$^{1}$,
    Tao Jiang$^{2}$,
    Hang Zhao$^{1,2}$
    \\[2ex]
    $^{1}$IIIS, Tsinghua University\\
    $^{2}$Galaxea AI
    \\
    {\tt\small yuanty22@mails.tsinghua.edu.cn}
}

\begin{document}

\maketitle
\thispagestyle{empty}
\pagestyle{empty}


\begin{strip}
    \centering
    \includegraphics[width=1.0\textwidth]{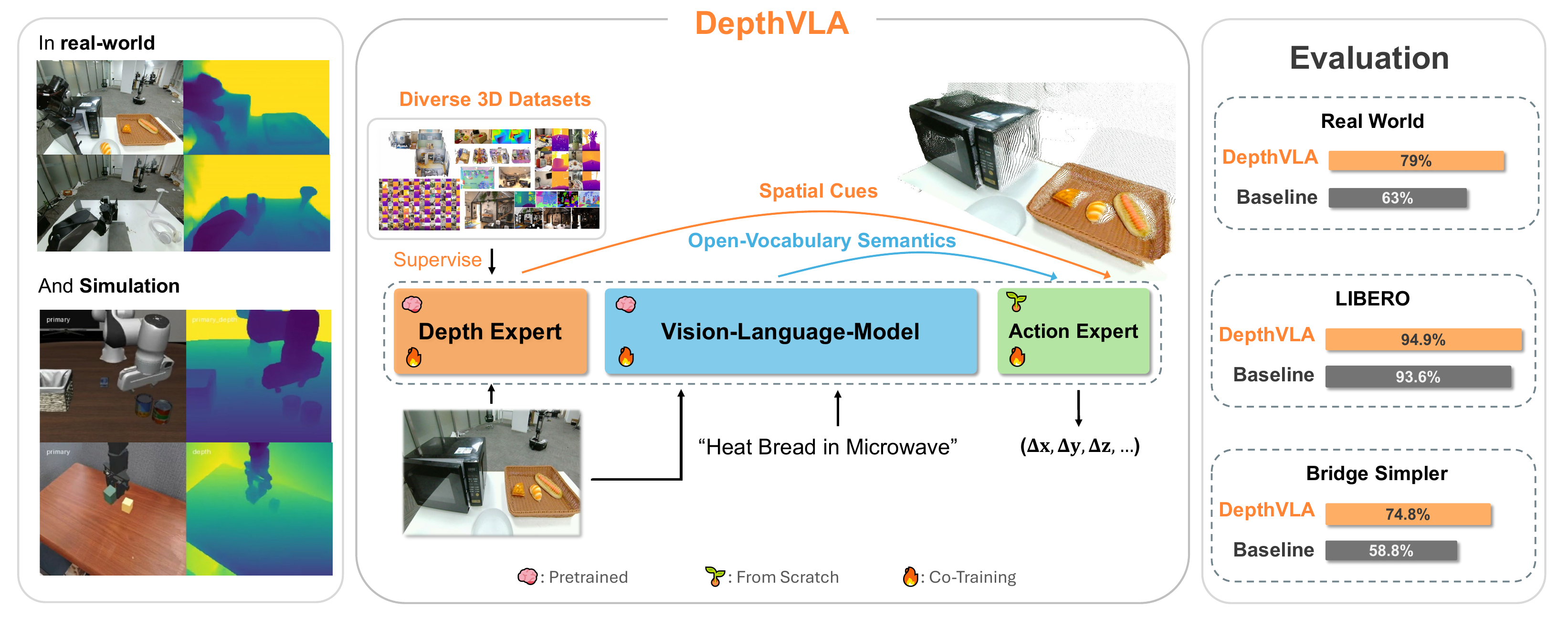}
    \captionof{figure}{We propose DepthVLA, a vision-language-action (VLA) model that explicitly incorporates spatial reasoning through a pretrained depth expert. Quantitative comparisons across Simpler, LIBERO, and real-world benchmarks show that DepthVLA consistently outperforms baselines, particularly in tasks requiring fine-grained 3D perception.}
    \label{fig:main_teaser}
\end{strip}

\begin{abstract}
Vision-Language-Action (VLA) models have recently shown impressive generalization and language-guided manipulation capabilities. However, their performance degrades on tasks requiring precise spatial reasoning due to limited spatial reasoning inherited from Vision-Language Models (VLMs). Existing VLAs rely on extensive action-data pretraining to ground VLMs in 3D space, which reduces training efficiency and is still insufficient for accurate spatial understanding.
In this work, we present \textbf{DepthVLA}, a simple yet effective VLA architecture that explicitly incorporates spatial awareness through a pretrained depth prediction module. DepthVLA adopts a mixture-of-transformers design that unifies a VLM, a depth transformer, and an action expert with fully shared attentions, forming an end-to-end model with enhanced spatial reasoning.
Extensive evaluations in both real-world and simulated environments show that DepthVLA outperforms state-of-the-art approaches, achieving 78.5\% vs. 65.0\% progress in real-world tasks, 94.9\% vs. 93.6\% in the LIBERO simulator, and 74.8\% vs. 58.8\% in the Simpler simulator. Our code will be made publicly available.

\end{abstract}

\section{INTRODUCTION}


Vision-Language-Action (VLA) models~\cite{kim24openvla, rt22023arxiv, black2024pi_0, octo_2023, li2024cogact, cheang2025gr3technicalreport} have emerged as a pivotal paradigm in robotic manipulation research. Built upon large-scale pretrained Vision-Language Models (VLMs), they inherit strong generalization capabilities from vast web data. VLMs provide robust language grounding and semantic visual perception, enabling VLAs to generalize across diverse tasks and embodiments. However, despite their strengths on semantics, VLMs exhibit limited spatial reasoning ability~\cite{Tong_2024_CVPR, Rahmanzadehgervi_2024_ACCV}, which in turn constrains the spatial perception abilities of VLAs, particularly in tasks requiring precise manipulation~\cite{qu2025spatialvla, li2025pointvlainjecting3dworld}. Current VLAs often rely on extensive action-data pretraining to ground VLMs in 3D space~\cite{kim24openvla, black2024pi_0, cheang2025gr3technicalreport, li2024cogact, rt22023arxiv, octo_2023}, which limits scalability, and pretrained VLAs continue
to struggle with precise spatial reasoning.
In practice, VLAs often fail at grasping small objects, executing precise operations, or avoiding collisions, highlighting their weak spatial perception.

Recent works have attempted to address this limitation by employing generative world models to predict future states~\cite{Zhao_2025_CVPR, zhen20243dvla, zhang2025up, zhu2025unifiedworldmodelscoupling, dreamvla25, tian2024predictive}. While promising, these methods lack explicit 3D knowledge, which we argue is essential for precise manipulation. Another line of work leverages Chain-of-Thought (CoT) reasoning~\cite{lee2025molmoactactionreasoningmodels} to autoregressively generate spatial tokens. However, this approach introduces significant latency (over 2 seconds), as hundreds of spatial tokens must be generated before action prediction. To overcome these limitations, we ask: \textbf{how can recent advances in 3D perception~\cite{depth_anything_v1, depth_anything_v2, wang2025vggt} be leveraged to enhance VLAs without sacrificing inference speed?}

To address this, we introduce \textbf{DepthVLA} (Figure~\ref{fig:main_teaser}), a simple yet effective VLA architecture that explicitly incorporates spatial awareness through a pretrained depth prediction expert. Trained on diverse 3D datasets~\cite{roberts:2021, xia2024rgbd, dai2017scannet, yeshwanth2023scannet++}, this module provides robust geometric understanding. Inspired by $\pi_0$~\cite{black2024pi_0}, DepthVLA uses a mixture-of-transformers (MoT)~\cite{liang2025mixtureoftransformers} design that integrates the depth expert with a VLM and a flow-matching action expert via fully shared attentions, forming an end-to-end VLA model. Intuitively, the VLM provides language understanding and open-vocabulary semantic perception, the depth expert provides fine-grained geometric cues, and the action expert generates actions conditioned on representations from both modalities. The MoT design also enables separate pretraining of each component, allowing training on a more diverse set of data beyond embodied action datasets. Despite adding a depth expert, DepthVLA only increases inference latency marginally, making it practical for real-time deployment.

We validate DepthVLA through extensive experiments in both real-world and simulated environments. We validate DepthVLA through extensive experiments in both real-world and simulated environments. Our evaluations show notable gains in grasping accuracy and collision avoidance, underscoring DepthVLA’s enhanced spatial reasoning. For real-world evaluation, we pretrain on the Galaxea Open-World Dataset~\cite{jiang2025galaxeaopenworlddatasetg0} and test on the Galaxea R1 Lite, a commercially available dual-arm mobile platform. In simulation, we evaluate on LIBERO~\cite{liu2023libero} and Simpler~\cite{li24simpler}. Results show that DepthVLA outperforms existing approaches, achieving 78.5\% vs. 65.0\% success in real-world tasks, 94.9\% vs. 93.6\% in LIBERO, and 74.8\% vs. 58.8\% in Simpler, demonstrating the effectiveness of depth-aware representations for precise, generalizable manipulation.

Our contributions are summarized as follows:

\begin{itemize}

\item \textbf{DepthVLA architecture: }We propose DepthVLA, a novel VLA model that integrates a pretrained depth prediction expert into a mixture-of-transformers framework, enabling explicit spatial reasoning while preserving semantic grounding from VLMs.
\item \textbf{Per-expert pretraining strategy:} Our MoT design allows each expert (VLM and depth) to be pretrained separately on diverse datasets, improving training efficiency and scalability beyond embodied action data.
\item \textbf{Extensive real-world and simulated validation:} We demonstrate that DepthVLA significantly outperforms state-of-the-art VLAs in both real-world and simulated environments (LIBERO, Simpler), achieving notable gains in grasping accuracy, collision avoidance, and overall task success.

\end{itemize}

\section{Related Work}
\subsection{Generalist Robot Manipulation Policies}
Robotic manipulation has evolved from single-task specialists to generalist models trained on broad, diverse datasets covering many tasks and embodiments. Fueled by advances in LLMs, VLMs~\cite{beyer2024paligemma, karamcheti2024prismaticvlmsinvestigatingdesign}, and large-scale robot action datasets~\cite{walke2023bridgedata, khazatsky2025droidlargescaleinthewildrobot}, this evolution has given rise to Vision-Language-Action (VLA) models. Early VLAs~\cite{kim24openvla, rt22023arxiv} typically fine-tuned VLMs to autoregressively generate action tokens, which facilitated knowledge transfer but incurred slow inference. More recent VLAs~\cite{piccinelli2025unidepthv2, cheang2025gr3technicalreport} adopt diffusion-based action experts to generate continuous actions more efficiently. Despite differences in action generation, most existing VLAs still require large-scale action-data pretraining to adapt to embodied settings, which is inefficient and still insufficient for
fine-grained spatial understanding.

\subsection{VLAs with Spatial Awareness}
Prior studies have shown that even state-of-the-art VLMs are insensitive to object shapes and fine geometry~\cite{Tong_2024_CVPR, Rahmanzadehgervi_2024_ACCV}, limiting their utility for precise manipulation. To enhance spatial perception, early efforts augmented VLAs with additional 3D inputs from sensors such as LiDAR or RGB-D cameras~\cite{li2025pointvlainjecting3dworld, li2025bridgevlainputoutputalignmentefficient, jia2024lift3dfoundationpolicylifting}, but this reduced generalizability across platforms. SpatialVLA~\cite{qu2025spatialvla} proposes using an off-the-shelf depth estimator to generate pseudo point clouds as input. However, this approach is essentially a workaround, as the depth estimator is not optimized end-to-end with the VLA, limiting its performance upper bound.

More recent approaches incorporate generative world models that predict future frames, keypoints, or semantic states, and then condition action generation on these predictions~\cite{Zhao_2025_CVPR, zhen20243dvla, zhang2025up, zhu2025unifiedworldmodelscoupling, dreamvla25}. While this improves planning by simulating futures, it does little to improve the encoding of the current scene. A concurrent line of work~\cite{lee2025molmoactactionreasoningmodels}, inspired by methods in VLMs~\cite{Bigverdi_2025_CVPR}, uses Chain-of-Thought (CoT) reasoning to autoregressively generate depth tokens. However, this strategy introduces high latency (over 2 seconds on modern GPUs), as hundreds of tokens must be auto-regressively generated before action prediction.

\subsection{3D Geometry Perception}
Recent advances in 3D perception~\cite{depth_anything_v1, depth_anything_v2, wang2025vggt, mast3r_eccv24, dust3r_cvpr24} have demonstrated strong ability to infer geometry from monocular or multi-view images. By scaling both 3D datasets and model capacity, these vision foundation models achieve robust spatial estimation and support downstream applications such as SLAM~\cite{zhang2024monst3r, murai2024_mast3rslam} and reconstruction~\cite{wang2024spann3r, long3r}. Their progress highlights the potential of integrating powerful 3D priors into VLAs for improved spatial reasoning without requiring additional sensors.

\section{Method}
\begin{figure*}[t]
  \centering
   \includegraphics[width=1.0\textwidth]{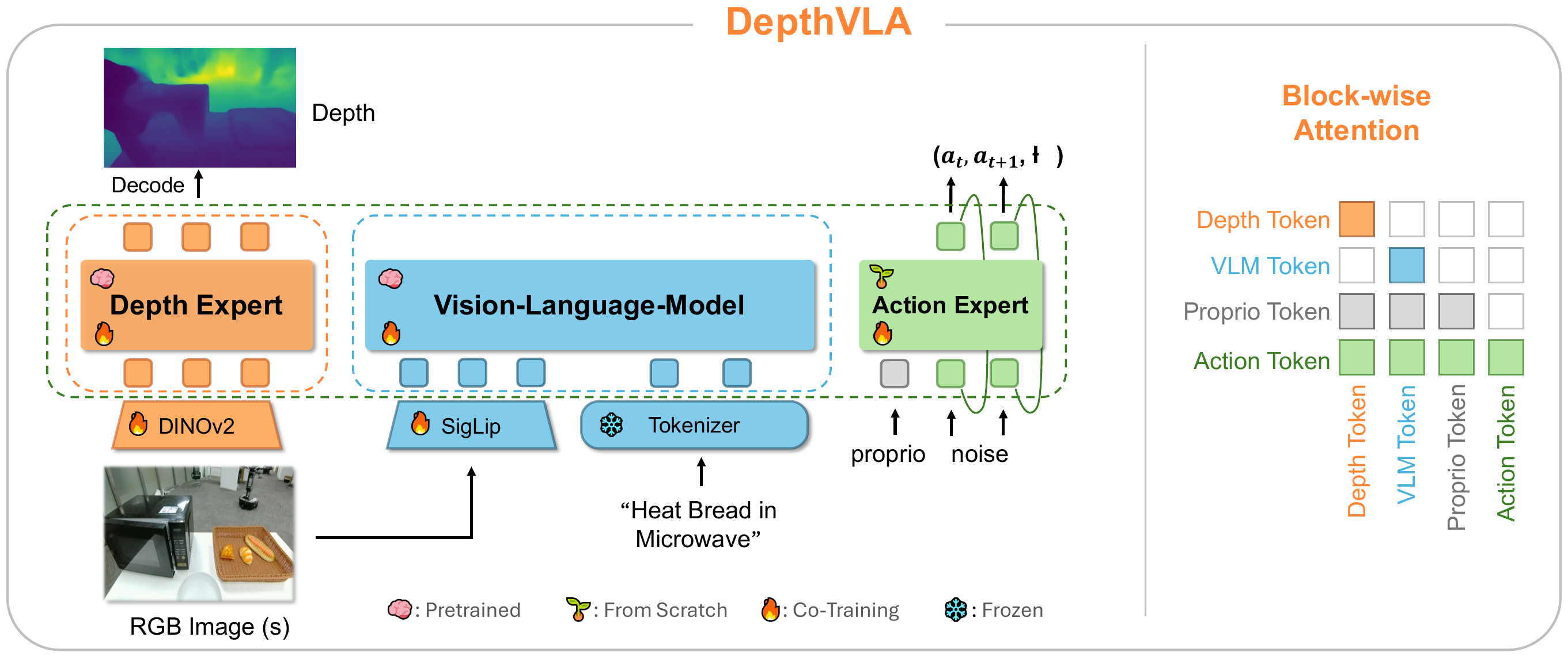}
   \caption{The proposed mixture-of-transformers (MoT) framework integrates three experts: a vision-language model (VLM) for semantic and language understanding, a depth expert for geometric reasoning, and an action expert for continuous control. Attention layers are shared across experts, while block-wise masking ensures pretrained modules retain their learned abilities. The action expert attends to features from both the VLM and depth expert at every layer to generate actions conditioned on language, visual, and spatial cues.}
   \label{fig:model}
\end{figure*}

In this section, we describe DepthVLA, its components, and the training framework.

\subsection{Problem Formulation and Model Overview}
We follow the standard end-to-end VLA setting, where a policy $\pi_{\theta}$ predicts a $k$-length action chunk 
$\boldsymbol{A_t}=a_{t:t+k}$
given the current observation $o_t$ (from one or multiple cameras), a language instruction $l$, and proprioceptive states $s_t$:,
$$\boldsymbol{A_t}=\pi_{\theta}\left(o_t, l, s_t\right).$$
DepthVLA adopts a mixture-of-transformers (MoT) architecture that integrates three experts: a VLM, a depth module, and a flow-matching action expert, as illustrated in Figure~\ref{fig:model}. This design extends $\pi_0$\cite{black2024pi_0}, which uses a two-expert MoT (VLM + action expert), by adding an independent depth expert to provide explicit spatial information.

Specifically, the VLM expert encodes $o_t$ and $l$ to capture semantic and language-grounded features, while the depth expert processes $o_t$ to infer geometric information. The action expert then generates continuous actions conditioned on the combined features from both semantic and geometric experts. All three experts share the same attention layers but maintain distinct weights and feature dimensions. 

To preserve the pretrained capabilities of the VLM and depth modules, we apply a block-wise mask: tokens from the VLM and depth experts attend only to themselves, while action tokens can attend to all streams, as shown in right side of Figure~\ref{fig:model}. This design allows DepthVLA to leverage pretrained knowledge while fusing semantic and spatial cues for precise action generation.

\subsection{Depth Expert}
The depth expert serves as a dedicated spatial reasoner, providing geometric cues to guide the action expert. To integrate seamlessly into the VLA, it adopts the same transformer backbone as the VLM, with separate weights and dimensions.

We design the depth expert as an encoder-decoder architecture. The encoder is based on DINOv2~\cite{oquab2023dinov2}, which captures fine-grained geometric features. We initialize from the pretrained checkpoint of Depth Anything V2~\cite{depth_anything_v2} to inherit strong spatial priors from large-scale 3D foundation models. The decoder mirrors the transformer structure of the VLM and outputs depth predictions through a linear head.
Unlike approaches that only provide a final depth map~\cite{dreamvla25, lee2025molmoactactionreasoningmodels}, we design the depth expert to perform spatial reasoning across all intermediate layers, which provides richer geometric cues for action prediction. The action expert attends to these intermediate features, leveraging rich geometric representations rather than low-dimensional depth outputs. This improves fine-grained spatial understanding, essential for tasks like precise grasping and collision avoidance.

Before integration to VLA, the depth expert is pretrained on diverse 3D datasets using a monocular depth prediction task to acquire robust spatial reasoning ability. We adopt the scale-invariant log loss~\cite{eigen2014depthmappredictionsingle}:
$$
\mathcal{L}_{\text{si}}(\hat{d}, d)=\sqrt{\frac{1}{n}\sum_i y^2 - \lambda \left(\frac{1}{n}\sum_i {y}\right)^2},
$$
$$
\text{where }y = \log\hat{d} - \log d.
$$
Here $d$ is the ground-truth metric depth, $\hat{d}$ is the predicted depth map, and $\lambda$ controls the balance of the scale term (set to 0.5 by default). This simple loss suffices for learning robust spatial reasoning and distance estimation.

\begin{table*}[t]
  \hspace{0.5em}
  \caption{Success rates on the Simpler WidowX benchmark. Models are trained on BridgeData V2 and evaluated zero-shot in simulation. The "Pretrained" column indicates whether the model is pretrained with additional robot action data. DepthVLA achieves the highest average performance.}
  \label{tab:bridge_main}
  \centering
  \resizebox{0.9\textwidth}{!}{
    \begin{tabular}{c|c|cccc|c}
    \toprule
    Model & Pretrained & Put Spoon & Put Carrot & Stack Block & Pick Eggplant & Average \\
    \midrule
    Diffusion Policy~\cite{chi2024diffusionpolicy} & $\times$ & 4.2\% & 0\% & 0\% & 0\% & 1.0\% \\ 
    Octo-Base~\cite{octo_2023} & \checkmark & 12.5\% & 8.3\% & 0\% & 43.1\% & 16.0\% \\
    SpatialVLA~\cite{qu2025spatialvla} & \checkmark  & 16.7\% & 25.0\% & 29.2\% & 100.0\% & 34.4\% \\
    $\pi_0$ (re-implemented)~\cite{black2024pi_0} & $\times$ & 81.7\% & 64.2\% & 30.0\% & 59.2\% & 58.8\%\\
    \textbf{DepthVLA (Ours)} & $\times$ & 75.8\% & 71.7\% & 62.5\% & 89.2\% & \textbf{74.8\%} \\
    \bottomrule
    \end{tabular}
  }
  
\end{table*}

\subsection{DepthVLA Policy Training}
We train DepthVLA on embodied action data with an imitation learning objective, maximizing the log-likelihood of actions:
$$
\max_{\theta} \; \mathbb{E}_{p(\boldsymbol{A}_{t}, {o}_t, l, {s}_t)} \left[ \log \pi_{\theta}(\boldsymbol{A}_{t} \mid {o}_t, l, {s}_t) \right]
$$
To better model continuous and diverse action trajectories, we adopt a flow-matching loss:
$$
\mathcal{L}_{\text{flow}}(\theta) = \mathbb{E}_{p(\boldsymbol{A}_t^\tau \mid {o}_t, l, {s}_t)} \left[
\left\|
v_{\theta}(\boldsymbol{A}_t^\tau, \tau, {o}_t, l, {s}_t)
-
\boldsymbol{u}(\boldsymbol{A}_t^\tau \mid \boldsymbol{A}_t)
\right\|^2
\right]
$$
Here, subscripts denote robot timesteps and superscripts denote flow matching timesteps, with $\tau \in [0, 1]$. $\boldsymbol{A}_t^\tau$ is the interpolated noisy action $\boldsymbol{A}_t^\tau=\tau\boldsymbol{A}_t+(1-\tau)\epsilon$. $v_{\theta}(\cdot)$ is the flow predicted by the model and $\boldsymbol{u}(\cdot)$ is the target flow derived from the action trajectory.

To maintain the depth expert’s spatial reasoning, we retain the depth prediction loss during the VLA training. The final loss is therefore:
$$
\mathcal{L}=\mathcal{L_\text{si}} + \mathcal{L}_{\text{flow}}.
$$
This approach allows DepthVLA to jointly optimize spatial reasoning and action generation in an end-to-end manner.

\section{Experiments}
\subsection{Implementation Details}
\noindent\textbf{Model Architecture.}
We implement all models in PyTorch. We use Paligemma-3B~\cite{beyer2024paligemma} as the VLM backbone, following prior VLA works~\cite{qu2025spatialvla, black2024pi_0, jiang2025galaxeaopenworlddatasetg0} due to its strong generalization ability. The depth expert employs DINOv2-L as the encoder, initialized from Depth Anything V2~\cite{depth_anything_v2}, while its decoder is matched in size to the action expert, with both modules containing approximately 300M parameters. 
As our closest baseline, we re-implement $\pi_0$ by strictly following the official JAX implementation. The only difference between our re-implemented $\pi_0$ and DepthVLA is the addition of the depth expert, allowing a fair comparison of the impact of explicit spatial reasoning.

\noindent\textbf{Training Details.}
The depth expert is pretrained on large-scale 3D datasets, including WildRGB-D~\cite{xia2024rgbd}, Scannet~\cite{dai2017scannet}, Scannet++~\cite{yeshwanth2023scannet++} and HyperSim~\cite{roberts:2021}. Pretraining runs for 50k steps using a cosine learning rate schedule, with batch size 1024 and initial learning rate $5 \times 10^{-5}$. 
For VLA training, we use a batch size of 1024 for large-scale datasets (e.g., Galaxea Open-World~\cite{jiang2025galaxeaopenworlddatasetg0}, BridgeData V2~\cite{walke2023bridgedata}) and 64 for smaller-scale datasets (e.g., LIBERO~\cite{liu2023libero}, real-world benchmark tasks). For all models, we do not use any historical information for action generation. All models are trained on 32 NVIDIA H100 GPUs with using the AdamW optimizer~\cite{loshchilov2019decoupledweightdecayregularization} with learning rate $2.5 \times 10^{-5}$ and weight decay $10^{-4}$.

\noindent\textbf{Inference Details.}
DepthVLA introduces 600M additional parameters compared with the baseline $\pi_0$ (300M from the DINOv2 encoder and 300M from the depth expert decoder). We run inference on an NVIDIA 4090 GPU with BF16 mixed precision. DepthVLA requires 8.0~GB of VRAM (vs. 6.7~GB for $\pi_0$) and has an inference latency of 210~ms per step (vs. 190~ms for $\pi_0$). Since actions are predicted in 1-second chunks (16 steps on a  15~Hz platform), the extra latency is negligible in practice.

\subsection{Simulation Benchmarking}
\noindent\textbf{BridgeV2 \& Simpler.} 
BridgeData V2~\cite{walke2023bridgedata} is a large-scale real-world robot manipulation dataset, containing over 60k trajectories collected across 24 environments using the WidowX robot. It provides diverse tasks and environment variations, making it a strong foundation for training generalist policies. To obtain depth supervision, we generate pseudo-labels using Depth Anything V2~\cite{depth_anything_v2} and UniDepth V2~\cite{piccinelli2025unidepthv2}.

Simpler WidowX~\cite{li24simpler} is a simulation environment designed to closely mirror BridgeData V2, providing a reproducible platform for policy evaluation. It includes four task suites with variations in environment, object configurations, and camera poses, effectively bridging the gap between real and simulated domains. We train DepthVLA on BridgeData V2 for 20k steps (approx. 12 epochs) and evaluate it zero-shot on Simpler WidowX. We report final success rate of each task suite, tested with 120 trials under different random seeds.

Results are shown in Table~\ref{tab:bridge_main}. The "Pretrained" column indicates whether a model was pretrained on additional robot action data. DepthVLA achieves the highest average success rate on Simpler WidowX. Compared with the counterpart without a depth expert (\ie, $\pi_0$ re-implemented), DepthVLA yields substantial gains on tasks such as block stacking and eggplant picking, which demand accurate spatial reasoning and collision avoidance. Remarkably, the depth expert improves 3D perception even when models are trained on real-world data but evaluated in simulation. Furthermore, DepthVLA outperforms SpatialVLA~\cite{qu2025spatialvla}, a spatial-aware VLA that leverages an external depth estimator, by a wide margin, highlighting the effectiveness of our mixture-of-transformers design.

\begin{table*}[t]
  \hspace{0.5em}
  \caption{Success rates on the LIBERO benchmark across four task suites. The "Pretrained" column indicates whether the model is pretrained with additional robot action data. DepthVLA outperforms all baselines, showing that explicit depth reasoning improves generalization across diverse manipulation tasks.}
  \label{tab:libero_main}
  \centering
  \begin{threeparttable}  
    \resizebox{0.8\textwidth}{!}{
      \begin{tabular}{c|c|cccc|c}
        \toprule
        Model & Pretrained & Spatial & Object & Goal & Long & Average \\
        \midrule
        Octo-Base~\cite{octo_2023} & \checkmark & 78.9\% & 85.7\% & 84.6\% & 51.1\% & 75.1\%\\
        OpenVLA~\cite{kim24openvla} & \checkmark & 84.7\% & 88.4\% & 79.2\% & 53.7\% & 76.5\% \\
        SpatialVLA~\cite{qu2025spatialvla}& \checkmark & 88.2\% & 89.9\% & 78.6\% & 55.5\% & 78.1\%\\
        CoT-VLA~\cite{Zhao_2025_CVPR} & \checkmark & 81.5\% & 91.6\% & 87.6\% & 69.0\% & 83.9\%\\
        MolmoACT~\cite{lee2025molmoactactionreasoningmodels} & \checkmark & 87.0\% & 95.4\% & 87.6\% & 77.2\% & 86.6\%\\
        DreamVLA~\cite{dreamvla25} & \checkmark & 97.5\% & 94.0\% & 89.5\% & 89.5\% & 92.6\%\\
        $\pi_0$ (re-implemented)~\cite{black2024pi_0} & $\times$ & 95.8\% & 96.4\% & 94.8\% & 87.4\% & 93.6\%\\
        $\pi_0$ (reported)~\cite{black2024pi_0}\tnote{*} & \checkmark & 96.8\% & 98.8\% & 95.8\% & 85.2\% & 94.2\%\\
        \textbf{DepthVLA (Ours)} & $\times$ &96.4\% & 98.0\% & 95.8\% & 89.2\% & \textbf{94.9\%}\\
        \bottomrule
      \end{tabular}
    }
      \noindent\parbox{0.8\textwidth}{
      \vspace{0.3em}\hspace{0.3em}
      \footnotesize
        \tnote{*} Reported in $\pi_0$ \href{https://github.com/Physical-Intelligence/openpi/blob/1f4506d6cd5b5a3188d753de8ccb2541f94f86c9/examples/libero/README.md}{official JAX implementation}.
      }
    
  \end{threeparttable}
\end{table*}
\noindent\textbf{LIBERO.}
LIBERO~\cite{liu2023libero} is a simulated manipulation benchmark based on the Franka Panda arm, with demonstrations that include front-view and wrist-view camera images along with natural language instructions. It comprises four task suites: LIBERO-Spatial/-Object/-Goal/-Long, each containing 500 demonstrations across 10 tasks.
Unlike prior works~\cite{kim24openvla, dreamvla25, black2024pi_0, lee2025molmoactactionreasoningmodels}, which typically train one model per suite, we train a single DepthVLA model jointly on all four suites for 30k steps (about 8 epochs). This creates a more challenging setting that requires stronger generalization across diverse task types. Success rates are reported per task suite, in total 2000 trials across 40 tasks with different random seeds.

Results are shown in Table~\ref{tab:libero_main}. The "Pretrained" column marks whether the model is pretrained on additional robot action datasets. DepthVLA achieves the highest average success rate, even surpassing all models with pretraining. This suggests that standard VLAs, even with large-scale action pretraining, still lack sufficient 3D grounding for precise manipulation. Moreover, DepthVLA surpasses both spatially enhanced baselines (\eg, MolmoACT~\cite{lee2025molmoactactionreasoningmodels}, SpatialVLA~\cite{qu2025spatialvla}) and world-model-based approaches (\eg, DreamVLA~\cite{dreamvla25}, CoTVLA~\cite{Zhao_2025_CVPR}), underscoring the strength of our depth expert design.

\subsection{Real-World Benchmarking}
\begin{figure}[th]
  \centering
   \includegraphics[width=0.8\linewidth]{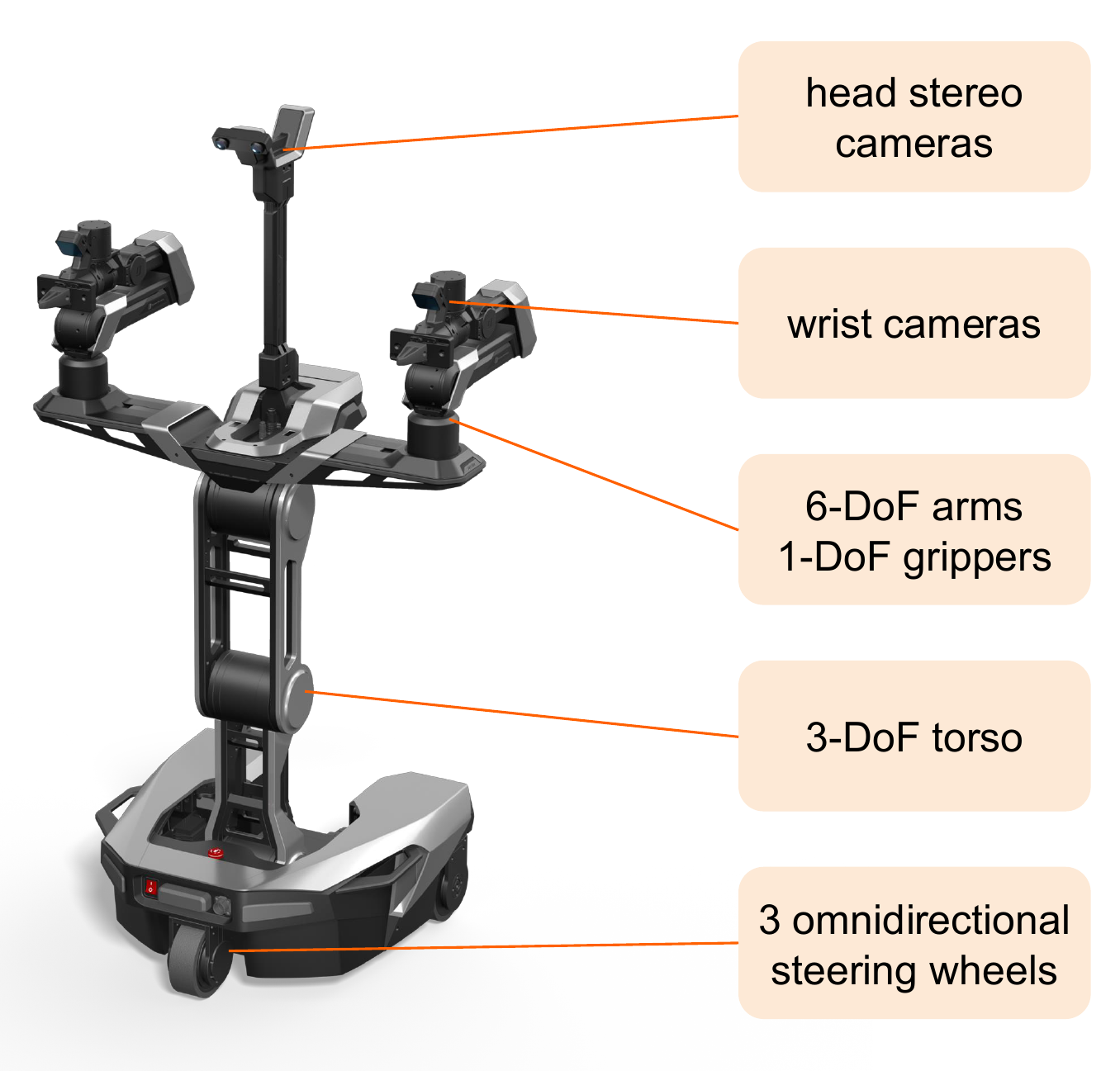}
   \caption{Real-robot experiment platform.}
   \label{fig:r1lite_platform}
\end{figure}

\begin{figure*}[t]
  \centering
   \includegraphics[width=1.0\textwidth]{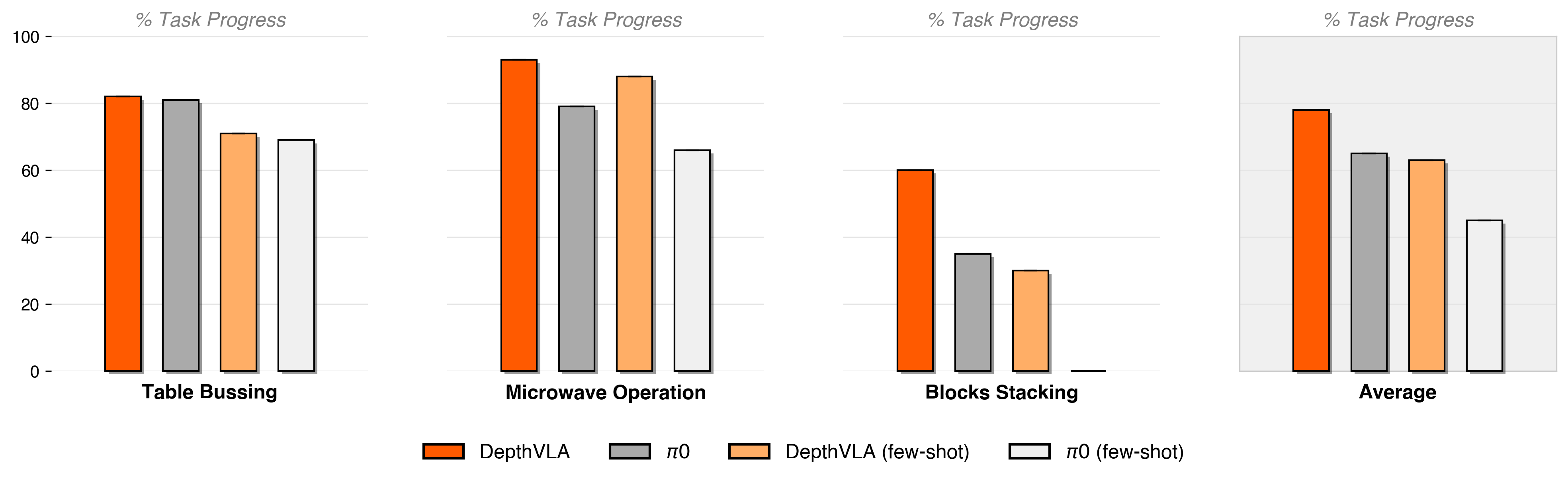}
   \caption{Performance of DepthVLA and baseline on three bimanual tasks with standard fine-tuning and few-shot adaptation. DepthVLA shows improvements in tasks requiring precise spatial reasoning and collision avoidance while maintaining comparable performance on simpler small-object manipulation.}
   \label{fig:real_world_results}
\end{figure*}

We evaluate DepthVLA on the Galaxea R1 Lite, a commercially available dual-arm mobile platform. The system consists of two 6-DoF arms, two wrist-mounted cameras, and a head camera, as shown in Figure~\ref{fig:r1lite_platform}. To assess the benefits of large-scale action pretraining on DepthVLA, we pretrain DepthVLA on the large-scale Galaxea Open-World Dataset~\cite{jiang2025galaxeaopenworlddatasetg0}, which contains 100k trajectories across 150 task categories and 50 real-world scenes. Depth labels are generated using VGGT~\cite{wang2025vggt} and UniDepth V2~\cite{piccinelli2025unidepthv2}. Pretraining runs for 80k steps (about 4 epochs) for both DepthVLA and the re-implemented $\pi_0$.

To evaluate spatial perception, fine-grained grasping, and collision avoidance, we design three benchmark tasks:

\noindent\textbf{Table bussing:} The robot organizes a cluttered desk by placing pens into a holder, hanging headphones, and moving a book onto a stand. This task measures small-object grasping and accurate position estimation.

\noindent\textbf{Microwave operation:} The robot opens a microwave door, places food on a plate, puts the plate inside, and closes the door. This task tests collision avoidance at each step.

\noindent\textbf{Blocks stacking:} The robot stacks blocks vertically, testing precise pick-and-place skills.

For each benchmark, we collect 100 trajectories and fine-tune the pretrained model for 4k steps. Performance is evaluated using progress scores, where each successful substep in a task contributes one point, and scores are averaged over 20 runs per task. Additionally, we also conduct few-shot experiments with only 20 fine-tuning trajectories to assess DepthVLA’s few-shot transferring ability.

Results are shown in Figure~\ref{fig:real_world_results}. DepthVLA consistently outperforms the baseline, achieving an average progress score of 79\% vs. 65\% in the standard fine-tuning setting, and 63\% vs. 45\% in the few-shot setting. On \textbf{microwave operation}, it demonstrates improved collision avoidance when handling the door and plate. On \textbf{block stacking}, DepthVLA exhibits superior spatial perception, even with limited fine-tuning data, whereas the baseline struggles. On \textbf{table bussing}, DepthVLA performs comparably to the baseline, suggesting that both models handle relatively simple small-object grasping tasks effectively. Importantly, DepthVLA maintains language-following capabilities, indicating that the action expert effectively integrates the strengths of both the VLM and depth expert.

\begin{figure*}[t]
  \centering
   \includegraphics[width=1.0\textwidth]{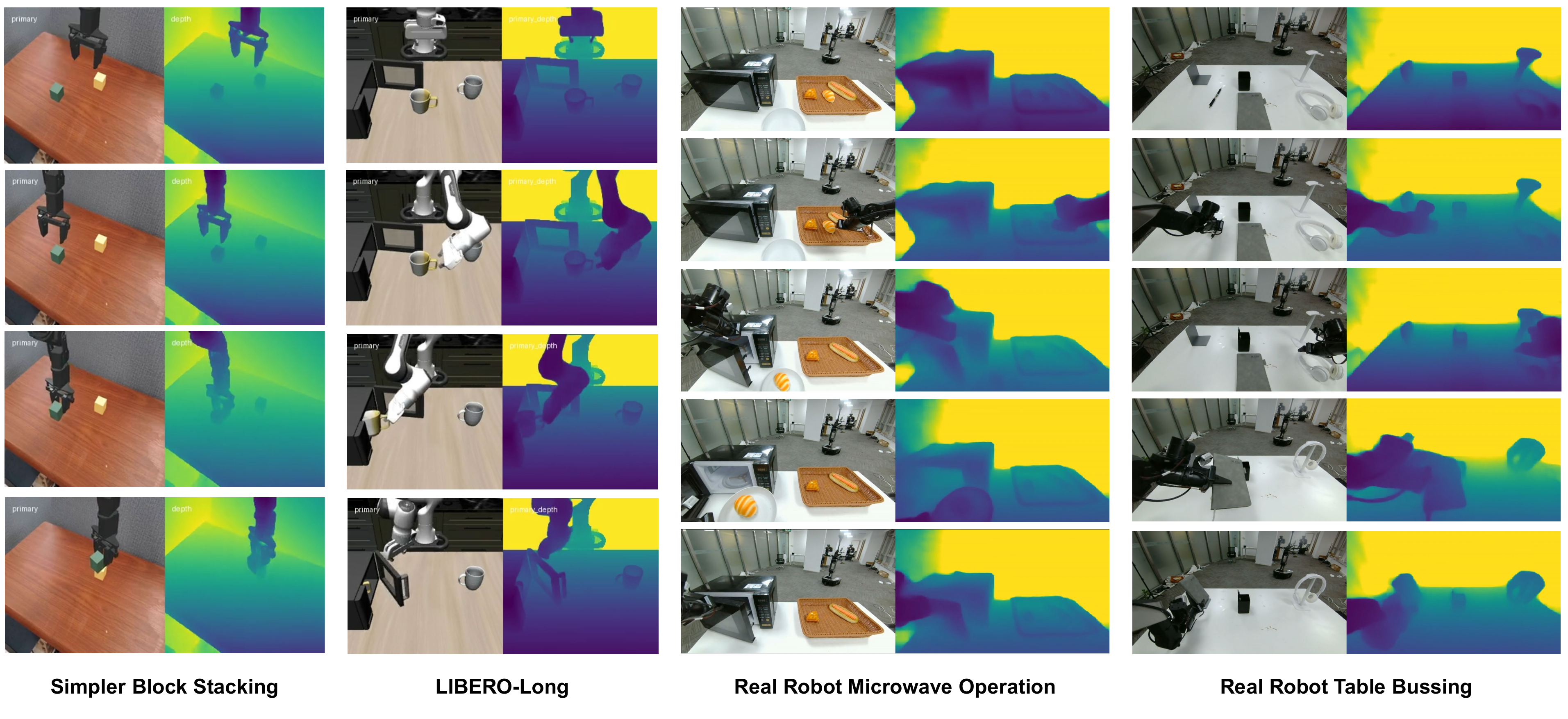}
   \caption{Qualitative results of DepthVLA’s predicted depth maps across real-world and simulated environments. The predicted depth provides fine-grained geometric cues that guide accurate manipulation, collision avoidance, and precise object grasping. Even in cluttered or challenging scenes, DepthVLA captures the 3D layout robustly, highlighting the effectiveness of the pretrained depth expert in providing spatial awareness.}
   \label{fig:exp_qualitative}
\end{figure*}

\subsection{Ablation Studies}
We conduct ablation studies to evaluate the design choices of the depth expert. Specifically, we investigate:
(i) Is pretraining the depth expert necessary?
(ii) Is the depth loss necessary during VLA training?
(iii) What happens if the depth expert is frozen during VLA training?
(iv) Is the block-wise mask between VLM and depth tokens necessary?
(v) Does predicting depth outperform directly inputting ground-truth depth?

We test these questions under the following settings:
(i) Depth expert randomly initialized without pretraining.
(ii) Depth loss removed during VLA training.
(iii) Depth expert frozen during VLA training.
(iv) Depth and VLM tokens allowed to attend to each other.
(v) Depth expert taking ground-truth depth as input.

Note that (ii) and (iii) differ, as the depth expert still receives gradients from the flow-matching loss in (ii). Settings (i)–(iv) are evaluated on BridgeData V2 \& Simpler, while (v) is evaluated on LIBERO, which provides ground-truth depth maps during inference.
\begin{table}[tb]
  \caption{Ablation studies on different design of the depth expert.}
  \centering
  \resizebox{1.0\linewidth}{!}{
    \begin{tabular}{c|cccc|c}
    \toprule
    Model & Spoon & Carrot & Block & Eggplant & Average \\
    \midrule
    (i) & 60.0\% & 60.8\% & 43.3\% & 40.0\% & 51.0\% \\ 
    (ii) & 69.2\% & 60\% & 28.3\% & 70.0\% & 56.9\% \\
    (iii) & 65.8\% & 69.2\% & 74.2\% & 78.3\% & 71.9\% \\ 
    (iv) & 66.7\% & 65.0\% & 2.5\% & 88.3\% & 55.6\% \\ 
    \textbf{DepthVLA} & 75.8\% & 71.7\% & 62.5\% & 89.2\% & \textbf{74.8\%} \\
    \bottomrule
    \end{tabular}
  }
  \label{tab:ablation_bridge}
\end{table}
\begin{table}[tb]
  \caption{Comparison between predicted and ground-truth depth inputs. Predicting depth yields stronger performance.}
  \centering
  \resizebox{1.0\linewidth}{!}{
    \begin{tabular}{c|cccc|c}
    \toprule
    Model & Spatial & Object & Goal & Long & Average \\
    \midrule
    (v) & 94.0\% & 97.6\% & 95.0\% & 86.4\% & 93.3\% \\ 
    \textbf{DepthVLA} &96.4\% & 98.0\% & 95.8\% & 89.2\% & \textbf{94.9\%}\\
    \bottomrule
    \end{tabular}
  }
  \label{tab:ablation_libero}
\end{table}

Results are summarized in Table~\ref{tab:ablation_bridge} and Table~\ref{tab:ablation_libero}. Each component proves essential for DepthVLA’s effectiveness. Notably, the performance is not greatly impacted when freezing the depth expert, which means the depth expert learned robust and universal spatial representation. It allows DepthVLA to be easily deployed by fine-tuning on demonstrations without the need of depth ground-truth.
Another interesting finding is that, the model performs better when predicting depth than when consuming ground-truth depth directly. We hypothesize this is due to modality competence~\cite{du2023unimodalfeaturelearningsupervised, Zhang_2024_CVPR_MultimodalRepresentation}, where one modality can dominate others when jointly provided. By learning to predict depth internally, DepthVLA avoids over-reliance on external signals and instead integrates geometric reasoning more effectively into the shared representation space.

\subsection{Visualization of Depth Prediction}
While DepthVLA primarily leverages intermediate features from the depth expert rather than its final outputs, we visualize predicted depth maps to better illustrate the model’s spatial perception capabilities.

As shown in Figure~\ref{fig:exp_qualitative}, the predicted depth captures detailed 3D structure, including object boundaries, distances, and occlusions, which are critical for precise manipulation. Notably, in cluttered environments such as the microwave operation, DepthVLA accurately estimates the relative positions of objects, supporting reliable grasping and collision avoidance. Similarly, in Simpler block stacking and LIBERO-long tasks, the depth predictions provide the action expert with fine-grained geometric cues that improve object alignment and positioning accuracy. 

These visualizations demonstrate that the depth expert effectively extracts 3D spatial information from monocular RGB input. This depth-aware representation complements the semantic grounding provided by the VLM and underpins the performance improvements observed across real and simulated environments.

\section{Conclusion}
We introduced DepthVLA, a VLA model that enhances spatial reasoning by integrating a pretrained depth expert with a VLM and action expert in a unified mixture-of-transformers framework. Experiments in both real-world and simulated environments show that DepthVLA improves performance on tasks requiring precise manipulation, collision avoidance, and fine-grained grasping, while preserving strong language-following capabilities. Ablations confirm the critical role of depth pretraining, depth loss, and attention design in achieving robust 3D perception.

Despite these improvements, DepthVLA has limitations: monocular depth prediction remains an ill-posed and challenging problem. Even when trained on diverse 3D datasets, the depth expert can struggle in difficult scenarios, such as tiny edges, reflective or transparent objects, or texture-less surfaces, which can impact action generation. Future work could explore multi-view depth or pointmap prediction~\cite{dust3r_cvpr24, mast3r_eccv24, wang2025vggt} to further enhance spatial accuracy and robustness.



\begin{thebibliography}{10}
\providecommand{\url}[1]{#1}
\csname url@rmstyle\endcsname
\providecommand{\newblock}{\relax}
\providecommand{\bibinfo}[2]{#2}
\providecommand\BIBentrySTDinterwordspacing{\spaceskip=0pt\relax}
\providecommand\BIBentryALTinterwordstretchfactor{4}
\providecommand\BIBentryALTinterwordspacing{\spaceskip=\fontdimen2\font plus
\BIBentryALTinterwordstretchfactor\fontdimen3\font minus \fontdimen4\font\relax}
\providecommand\BIBforeignlanguage[2]{{%
\expandafter\ifx\csname l@#1\endcsname\relax
\typeout{** WARNING: IEEEtran.bst: No hyphenation pattern has been}%
\typeout{** loaded for the language `#1'. Using the pattern for}%
\typeout{** the default language instead.}%
\else
\language=\csname l@#1\endcsname
\fi
#2}}

\bibitem{kim24openvla}
M.~Kim, K.~Pertsch, S.~Karamcheti, T.~Xiao, A.~Balakrishna, S.~Nair, R.~Rafailov, E.~Foster, G.~Lam, P.~Sanketi, Q.~Vuong, T.~Kollar, B.~Burchfiel, R.~Tedrake, D.~Sadigh, S.~Levine, P.~Liang, and C.~Finn, ``Openvla: An open-source vision-language-action model,'' \emph{arXiv preprint arXiv:2406.09246}, 2024.

\bibitem{rt22023arxiv}
A.~Brohan, N.~Brown, J.~Carbajal, Y.~Chebotar, X.~Chen, K.~Choromanski, T.~Ding, D.~Driess, A.~Dubey, C.~Finn, P.~Florence, C.~Fu, M.~G. Arenas, K.~Gopalakrishnan, K.~Han, K.~Hausman, A.~Herzog, J.~Hsu, B.~Ichter, A.~Irpan, N.~Joshi, R.~Julian, D.~Kalashnikov, Y.~Kuang, I.~Leal, L.~Lee, T.-W.~E. Lee, S.~Levine, Y.~Lu, H.~Michalewski, I.~Mordatch, K.~Pertsch, K.~Rao, K.~Reymann, M.~Ryoo, G.~Salazar, P.~Sanketi, P.~Sermanet, J.~Singh, A.~Singh, R.~Soricut, H.~Tran, V.~Vanhoucke, Q.~Vuong, A.~Wahid, S.~Welker, P.~Wohlhart, J.~Wu, F.~Xia, T.~Xiao, P.~Xu, S.~Xu, T.~Yu, and B.~Zitkovich, ``Rt-2: Vision-language-action models transfer web knowledge to robotic control,'' in \emph{arXiv preprint arXiv:2307.15818}, 2023.

\bibitem{black2024pi_0}
K.~Black, N.~Brown, D.~Driess, A.~Esmail, M.~Equi, C.~Finn, N.~Fusai, L.~Groom, K.~Hausman, B.~Ichter, \emph{et~al.}, ``$\pi_0$: A vision-language-action flow model for general robot control,'' \emph{arXiv preprint arXiv:2410.24164}, 2024.

\bibitem{octo_2023}
{Octo Model Team}, D.~Ghosh, H.~Walke, K.~Pertsch, K.~Black, O.~Mees, S.~Dasari, J.~Hejna, C.~Xu, J.~Luo, T.~Kreiman, Y.~Tan, L.~Y. Chen, P.~Sanketi, Q.~Vuong, T.~Xiao, D.~Sadigh, C.~Finn, and S.~Levine, ``Octo: An open-source generalist robot policy,'' in \emph{Proceedings of Robotics: Science and Systems}, Delft, Netherlands, 2024.

\bibitem{li2024cogact}
Q.~Li, Y.~Liang, Z.~Wang, L.~Luo, X.~Chen, M.~Liao, F.~Wei, Y.~Deng, S.~Xu, Y.~Zhang, \emph{et~al.}, ``Cogact: A foundational vision-language-action model for synergizing cognition and action in robotic manipulation,'' \emph{arXiv preprint arXiv:2411.19650}, 2024.

\bibitem{cheang2025gr3technicalreport}
\BIBentryALTinterwordspacing
C.~Cheang, S.~Chen, Z.~Cui, Y.~Hu, L.~Huang, T.~Kong, H.~Li, Y.~Li, Y.~Liu, X.~Ma, H.~Niu, W.~Ou, W.~Peng, Z.~Ren, H.~Shi, J.~Tian, H.~Wu, X.~Xiao, Y.~Xiao, J.~Xu, and Y.~Yang, ``Gr-3 technical report,'' 2025. [Online]. Available: \url{https://arxiv.org/abs/2507.15493}
\BIBentrySTDinterwordspacing

\bibitem{Tong_2024_CVPR}
S.~Tong, Z.~Liu, Y.~Zhai, Y.~Ma, Y.~LeCun, and S.~Xie, ``Eyes wide shut? exploring the visual shortcomings of multimodal llms,'' in \emph{Proceedings of the IEEE/CVF Conference on Computer Vision and Pattern Recognition (CVPR)}, June 2024, pp. 9568--9578.

\bibitem{Rahmanzadehgervi_2024_ACCV}
P.~Rahmanzadehgervi, L.~Bolton, M.~R. Taesiri, and A.~T. Nguyen, ``Vision language models are blind,'' in \emph{Proceedings of the Asian Conference on Computer Vision (ACCV)}, December 2024, pp. 18--34.

\bibitem{qu2025spatialvla}
D.~Qu, H.~Song, Q.~Chen, Y.~Yao, X.~Ye, Y.~Ding, Z.~Wang, J.~Gu, B.~Zhao, D.~Wang, \emph{et~al.}, ``Spatialvla: Exploring spatial representations for visual-language-action model,'' \emph{arXiv preprint arXiv:2501.15830}, 2025.

\bibitem{li2025pointvlainjecting3dworld}
\BIBentryALTinterwordspacing
C.~Li, J.~Wen, Y.~Peng, Y.~Peng, F.~Feng, and Y.~Zhu, ``Pointvla: Injecting the 3d world into vision-language-action models,'' 2025. [Online]. Available: \url{https://arxiv.org/abs/2503.07511}
\BIBentrySTDinterwordspacing

\bibitem{Zhao_2025_CVPR}
Q.~Zhao, Y.~Lu, M.~J. Kim, Z.~Fu, Z.~Zhang, Y.~Wu, Z.~Li, Q.~Ma, S.~Han, C.~Finn, A.~Handa, T.-Y. Lin, G.~Wetzstein, M.-Y. Liu, and D.~Xiang, ``Cot-vla: Visual chain-of-thought reasoning for vision-language-action models,'' in \emph{Proceedings of the IEEE/CVF Conference on Computer Vision and Pattern Recognition (CVPR)}, June 2025, pp. 1702--1713.

\bibitem{zhen20243dvla}
H.~Zhen, X.~Qiu, P.~Chen, J.~Yang, X.~Yan, Y.~Du, Y.~Hong, and C.~Gan, ``3d-vla: 3d vision-language-action generative world model,'' \emph{arXiv preprint arXiv:2403.09631}, 2024.

\bibitem{zhang2025up}
J.~Zhang, Y.~Guo, Y.~Hu, X.~Chen, X.~Zhu, and J.~Chen, ``Up-vla: A unified understanding and prediction model for embodied agent,'' \emph{arXiv preprint arXiv:2501.18867}, 2025.

\bibitem{zhu2025unifiedworldmodelscoupling}
\BIBentryALTinterwordspacing
C.~Zhu, R.~Yu, S.~Feng, B.~Burchfiel, P.~Shah, and A.~Gupta, ``Unified world models: Coupling video and action diffusion for pretraining on large robotic datasets,'' 2025. [Online]. Available: \url{https://arxiv.org/abs/2504.02792}
\BIBentrySTDinterwordspacing

\bibitem{dreamvla25}
\BIBentryALTinterwordspacing
W.~Zhang, H.~Liu, Z.~Qi, Y.~Wang, X.~Yu, J.~Zhang, R.~Dong, J.~He, H.~Wang, Z.~Zhang, L.~Yi, W.~Zeng, and X.~Jin, ``Dreamvla: A vision-language-action model dreamed with comprehensive world knowledge,'' \emph{CoRR}, vol. abs/2507.04447, 2025. [Online]. Available: \url{https://doi.org/10.48550/arXiv.2507.04447}
\BIBentrySTDinterwordspacing

\bibitem{tian2024predictive}
Y.~Tian, S.~Yang, J.~Zeng, P.~Wang, D.~Lin, H.~Dong, and J.~Pang, ``Predictive inverse dynamics models are scalable learners for robotic manipulation,'' \emph{arXiv preprint arXiv:2412.15109}, 2024.

\bibitem{lee2025molmoactactionreasoningmodels}
\BIBentryALTinterwordspacing
J.~Lee, J.~Duan, H.~Fang, Y.~Deng, S.~Liu, B.~Li, B.~Fang, J.~Zhang, Y.~R. Wang, S.~Lee, W.~Han, W.~Pumacay, A.~Wu, R.~Hendrix, K.~Farley, E.~VanderBilt, A.~Farhadi, D.~Fox, and R.~Krishna, ``Molmoact: Action reasoning models that can reason in space,'' 2025. [Online]. Available: \url{https://arxiv.org/abs/2508.07917}
\BIBentrySTDinterwordspacing

\bibitem{depth_anything_v1}
L.~Yang, B.~Kang, Z.~Huang, X.~Xu, J.~Feng, and H.~Zhao, ``Depth anything: Unleashing the power of large-scale unlabeled data,'' in \emph{CVPR}, 2024.

\bibitem{depth_anything_v2}
L.~Yang, B.~Kang, Z.~Huang, Z.~Zhao, X.~Xu, J.~Feng, and H.~Zhao, ``Depth anything v2,'' \emph{arXiv:2406.09414}, 2024.

\bibitem{wang2025vggt}
J.~Wang, M.~Chen, N.~Karaev, A.~Vedaldi, C.~Rupprecht, and D.~Novotny, ``Vggt: Visual geometry grounded transformer,'' in \emph{Proceedings of the IEEE/CVF Conference on Computer Vision and Pattern Recognition}, 2025.

\bibitem{roberts:2021}
M.~Roberts, J.~Ramapuram, A.~Ranjan, A.~Kumar, M.~A. Bautista, N.~Paczan, R.~Webb, and J.~M. Susskind, ``{Hypersim}: {A} photorealistic synthetic dataset for holistic indoor scene understanding,'' in \emph{International Conference on Computer Vision (ICCV) 2021}, 2021.

\bibitem{xia2024rgbd}
H.~Xia, Y.~Fu, S.~Liu, and X.~Wang, ``Rgbd objects in the wild: Scaling real-world 3d object learning from rgb-d videos,'' 2024.

\bibitem{dai2017scannet}
A.~Dai, A.~X. Chang, M.~Savva, M.~Halber, T.~Funkhouser, and M.~Nie{\ss}ner, ``Scannet: Richly-annotated 3d reconstructions of indoor scenes,'' in \emph{Proc. Computer Vision and Pattern Recognition (CVPR), IEEE}, 2017.

\bibitem{yeshwanth2023scannet++}
C.~Yeshwanth, Y.-C. Liu, M.~Nie{\ss}ner, and A.~Dai, ``Scannet++: A high-fidelity dataset of 3d indoor scenes,'' in \emph{Proceedings of the IEEE/CVF International Conference on Computer Vision}, 2023, pp. 12--22.

\bibitem{liang2025mixtureoftransformers}
\BIBentryALTinterwordspacing
W.~Liang, L.~YU, L.~Luo, S.~Iyer, N.~Dong, C.~Zhou, G.~Ghosh, M.~Lewis, W.~tau Yih, L.~Zettlemoyer, and X.~V. Lin, ``Mixture-of-transformers: A sparse and scalable architecture for multi-modal foundation models,'' \emph{Transactions on Machine Learning Research}, 2025. [Online]. Available: \url{https://openreview.net/forum?id=Nu6N69i8SB}
\BIBentrySTDinterwordspacing

\bibitem{jiang2025galaxeaopenworlddatasetg0}
\BIBentryALTinterwordspacing
T.~Jiang, T.~Yuan, Y.~Liu, C.~Lu, J.~Cui, X.~Liu, S.~Cheng, J.~Gao, H.~Xu, and H.~Zhao, ``Galaxea open-world dataset and g0 dual-system vla model,'' 2025. [Online]. Available: \url{https://arxiv.org/abs/2509.00576}
\BIBentrySTDinterwordspacing

\bibitem{liu2023libero}
B.~Liu, Y.~Zhu, C.~Gao, Y.~Feng, Q.~Liu, Y.~Zhu, and P.~Stone, ``Libero: Benchmarking knowledge transfer for lifelong robot learning,'' \emph{arXiv preprint arXiv:2306.03310}, 2023.

\bibitem{li24simpler}
X.~Li, K.~Hsu, J.~Gu, K.~Pertsch, O.~Mees, H.~R. Walke, C.~Fu, I.~Lunawat, I.~Sieh, S.~Kirmani, S.~Levine, J.~Wu, C.~Finn, H.~Su, Q.~Vuong, and T.~Xiao, ``Evaluating real-world robot manipulation policies in simulation,'' \emph{arXiv preprint arXiv:2405.05941}, 2024.

\bibitem{beyer2024paligemma}
L.~Beyer, A.~Steiner, A.~S. Pinto, A.~Kolesnikov, X.~Wang, D.~Salz, M.~Neumann, I.~Alabdulmohsin, M.~Tschannen, E.~Bugliarello, T.~Unterthiner, D.~Keysers, S.~Koppula, F.~Liu, A.~Grycner, A.~Gritsenko, N.~Houlsby, M.~Kumar, K.~Rong, J.~Eisenschlos, R.~Kabra, M.~Bauer, M.~Bošnjak, X.~Chen, M.~Minderer, P.~Voigtlaender, I.~Bica, I.~Balazevic, J.~Puigcerver, P.~Papalampidi, O.~Henaff, X.~Xiong, R.~Soricut, J.~Harmsen, and X.~Zhai, ``{PaliGemma: A versatile 3B VLM for transfer},'' \emph{arXiv preprint arXiv:2407.07726}, 2024.

\bibitem{karamcheti2024prismaticvlmsinvestigatingdesign}
\BIBentryALTinterwordspacing
S.~Karamcheti, S.~Nair, A.~Balakrishna, P.~Liang, T.~Kollar, and D.~Sadigh, ``Prismatic vlms: Investigating the design space of visually-conditioned language models,'' 2024. [Online]. Available: \url{https://arxiv.org/abs/2402.07865}
\BIBentrySTDinterwordspacing

\bibitem{walke2023bridgedata}
H.~Walke, K.~Black, A.~Lee, M.~J. Kim, M.~Du, C.~Zheng, T.~Zhao, P.~Hansen-Estruch, Q.~Vuong, A.~He, V.~Myers, K.~Fang, C.~Finn, and S.~Levine, ``Bridgedata v2: A dataset for robot learning at scale,'' in \emph{Conference on Robot Learning (CoRL)}, 2023.

\bibitem{khazatsky2025droidlargescaleinthewildrobot}
\BIBentryALTinterwordspacing
A.~Khazatsky, K.~Pertsch, S.~Nair, A.~Balakrishna, S.~Dasari, S.~Karamcheti, S.~Nasiriany, M.~K. Srirama, L.~Y. Chen, K.~Ellis, P.~D. Fagan, J.~Hejna, M.~Itkina, M.~Lepert, Y.~J. Ma, P.~T. Miller, J.~Wu, S.~Belkhale, S.~Dass, H.~Ha, A.~Jain, A.~Lee, Y.~Lee, M.~Memmel, S.~Park, I.~Radosavovic, K.~Wang, A.~Zhan, K.~Black, C.~Chi, K.~B. Hatch, S.~Lin, J.~Lu, J.~Mercat, A.~Rehman, P.~R. Sanketi, A.~Sharma, C.~Simpson, Q.~Vuong, H.~R. Walke, B.~Wulfe, T.~Xiao, J.~H. Yang, A.~Yavary, T.~Z. Zhao, C.~Agia, R.~Baijal, M.~G. Castro, D.~Chen, Q.~Chen, T.~Chung, J.~Drake, E.~P. Foster, J.~Gao, V.~Guizilini, D.~A. Herrera, M.~Heo, K.~Hsu, J.~Hu, M.~Z. Irshad, D.~Jackson, C.~Le, Y.~Li, K.~Lin, R.~Lin, Z.~Ma, A.~Maddukuri, S.~Mirchandani, D.~Morton, T.~Nguyen, A.~O'Neill, R.~Scalise, D.~Seale, V.~Son, S.~Tian, E.~Tran, A.~E. Wang, Y.~Wu, A.~Xie, J.~Yang, P.~Yin, Y.~Zhang, O.~Bastani, G.~Berseth, J.~Bohg, K.~Goldberg, A.~Gupta, A.~Gupta, D.~Jayaraman, J.~J. Lim, J.~Malik, R.~Martín-Martín, S.~Ramamoorthy, D.~Sadigh,
  S.~Song, J.~Wu, M.~C. Yip, Y.~Zhu, T.~Kollar, S.~Levine, and C.~Finn, ``Droid: A large-scale in-the-wild robot manipulation dataset,'' 2025. [Online]. Available: \url{https://arxiv.org/abs/2403.12945}
\BIBentrySTDinterwordspacing

\bibitem{piccinelli2025unidepthv2}
\BIBentryALTinterwordspacing
L.~Piccinelli, C.~Sakaridis, Y.-H. Yang, M.~Segu, S.~Li, W.~Abbeloos, and L.~V. Gool, ``{U}ni{D}epth{V2}: Universal monocular metric depth estimation made simpler,'' 2025. [Online]. Available: \url{https://arxiv.org/abs/2502.20110}
\BIBentrySTDinterwordspacing

\bibitem{li2025bridgevlainputoutputalignmentefficient}
\BIBentryALTinterwordspacing
P.~Li, Y.~Chen, H.~Wu, X.~Ma, X.~Wu, Y.~Huang, L.~Wang, T.~Kong, and T.~Tan, ``Bridgevla: Input-output alignment for efficient 3d manipulation learning with vision-language models,'' 2025. [Online]. Available: \url{https://arxiv.org/abs/2506.07961}
\BIBentrySTDinterwordspacing

\bibitem{jia2024lift3dfoundationpolicylifting}
\BIBentryALTinterwordspacing
Y.~Jia, J.~Liu, S.~Chen, C.~Gu, Z.~Wang, L.~Luo, L.~Lee, P.~Wang, Z.~Wang, R.~Zhang, and S.~Zhang, ``Lift3d foundation policy: Lifting 2d large-scale pretrained models for robust 3d robotic manipulation,'' 2024. [Online]. Available: \url{https://arxiv.org/abs/2411.18623}
\BIBentrySTDinterwordspacing

\bibitem{Bigverdi_2025_CVPR}
M.~Bigverdi, Z.~Luo, C.-Y. Hsieh, E.~Shen, D.~Chen, L.~G. Shapiro, and R.~Krishna, ``Perception tokens enhance visual reasoning in multimodal language models,'' in \emph{Proceedings of the IEEE/CVF Conference on Computer Vision and Pattern Recognition (CVPR)}, June 2025, pp. 3836--3845.

\bibitem{mast3r_eccv24}
V.~Leroy, Y.~Cabon, and J.~Revaud, ``Grounding image matching in 3d with mast3r,'' 2024.

\bibitem{dust3r_cvpr24}
S.~Wang, V.~Leroy, Y.~Cabon, B.~Chidlovskii, and J.~Revaud, ``Dust3r: Geometric 3d vision made easy,'' in \emph{CVPR}, 2024.

\bibitem{zhang2024monst3r}
J.~Zhang, C.~Herrmann, J.~Hur, V.~Jampani, T.~Darrell, F.~Cole, D.~Sun, and M.-H. Yang, ``Monst3r: A simple approach for estimating geometry in the presence of motion,'' \emph{arXiv preprint arxiv:2410.03825}, 2024.

\bibitem{murai2024_mast3rslam}
R.~Murai, E.~Dexheimer, and A.~J. Davison, ``{MASt3R-SLAM}: Real-time dense {SLAM} with {3D} reconstruction priors,'' \emph{arXiv preprint}, 2024.

\bibitem{wang2024spann3r}
H.~Wang and L.~Agapito, ``3d reconstruction with spatial memory,'' \emph{arXiv preprint arXiv:2408.16061}, 2024.

\bibitem{long3r}
Z.~Chen, M.~Qin, T.~Yuan, Z.~Liu, and H.~Zhao, ``Long3r: Long sequence streaming 3d reconstruction,'' \emph{arXiv preprint arXiv:2507.18255}, 2025.

\bibitem{oquab2023dinov2}
M.~Oquab, T.~Darcet, T.~Moutakanni, H.~V. Vo, M.~Szafraniec, V.~Khalidov, P.~Fernandez, D.~Haziza, F.~Massa, A.~El-Nouby, R.~Howes, P.-Y. Huang, H.~Xu, V.~Sharma, S.-W. Li, W.~Galuba, M.~Rabbat, M.~Assran, N.~Ballas, G.~Synnaeve, I.~Misra, H.~Jegou, J.~Mairal, P.~Labatut, A.~Joulin, and P.~Bojanowski, ``Dinov2: Learning robust visual features without supervision,'' 2023.

\bibitem{eigen2014depthmappredictionsingle}
\BIBentryALTinterwordspacing
D.~Eigen, C.~Puhrsch, and R.~Fergus, ``Depth map prediction from a single image using a multi-scale deep network,'' 2014. [Online]. Available: \url{https://arxiv.org/abs/1406.2283}
\BIBentrySTDinterwordspacing

\bibitem{chi2024diffusionpolicy}
C.~Chi, Z.~Xu, S.~Feng, E.~Cousineau, Y.~Du, B.~Burchfiel, R.~Tedrake, and S.~Song, ``Diffusion policy: Visuomotor policy learning via action diffusion,'' \emph{The International Journal of Robotics Research}, 2024.

\bibitem{loshchilov2019decoupledweightdecayregularization}
\BIBentryALTinterwordspacing
I.~Loshchilov and F.~Hutter, ``Decoupled weight decay regularization,'' 2019. [Online]. Available: \url{https://arxiv.org/abs/1711.05101}
\BIBentrySTDinterwordspacing

\bibitem{du2023unimodalfeaturelearningsupervised}
\BIBentryALTinterwordspacing
C.~Du, J.~Teng, T.~Li, Y.~Liu, T.~Yuan, Y.~Wang, Y.~Yuan, and H.~Zhao, ``On uni-modal feature learning in supervised multi-modal learning,'' 2023. [Online]. Available: \url{https://arxiv.org/abs/2305.01233}
\BIBentrySTDinterwordspacing

\bibitem{Zhang_2024_CVPR_MultimodalRepresentation}
X.~Zhang, J.~Yoon, M.~Bansal, and H.~Yao, ``Multimodal representation learning by alternating unimodal adaptation,'' in \emph{Proceedings of the IEEE/CVF Conference on Computer Vision and Pattern Recognition (CVPR)}, June 2024, pp. 27\,456--27\,466.

\end{thebibliography}








\end{document}